\documentclass{article}

\IfFileExists{neurips_2022.sty}{
  \usepackage[preprint]{neurips_2022}
}{
  \usepackage[margin=1in]{geometry}
  \usepackage[numbers,sort&compress]{natbib}
}

\usepackage[utf8]{inputenc}
\usepackage[T1]{fontenc}
\usepackage{hyperref}
\usepackage{url}
\usepackage{booktabs}
\usepackage{amsmath}
\usepackage{amsfonts}
\usepackage{nicefrac}
\usepackage{microtype}
\usepackage[table]{xcolor}
\usepackage{graphicx}
\usepackage{svg}
\usepackage{float}
\usepackage{array}
\usepackage{enumitem}
\usepackage{caption}
\usepackage{placeins}
\usepackage{tabularx}
\usepackage{array}

\graphicspath{{figures/}}
\setlist{nosep,leftmargin=*}
\title{Edu-QuRating: Multi-Dimensional Educational Data Curation with Distilled Pairwise Judgements}

\author{
    \textbf{Oliver G. B. Garrod$^{1}$} \and \textbf{Robin A. A. Ince$^{1}$} \and \textbf{Meng Liu$^{1}$} \and \textbf{Mohamed Huti$^{1}$} \and \textbf{Moritz Boos$^{1}$} \and \textbf{Amy Waldock$^{1}$} \and \textbf{Dominic Andrews$^{1}$} \and \textbf{Romana Alonso-Kropil$^{1}$} \and \textbf{Paul Atherton$^{1}$} \\[20pt]
    {\normalsize $^1$Fab AI}
}

\begin{document}

\maketitle

\begin{abstract}
Educational data filters have become a practical way to improve language-model pre-training, but most filters treat educational value as a single scalar property. This may be too broad for some applications, especially if the data set already features a high density of educational material. Useful learning material needs to be accurate, engaging, well structured, and appropriate for the intended audience and application (e.g. learner- vs teacher-facing). Following QuRating \citep{wettig2024qurating}, we introduce Edu-QuRating: a pipeline for multi-dimensional educational data scoring and curation. Edu-QuRating defines education-specific rubrics, uses an LLM judge to label sampled document pairs and distills those pairwise preferences into reusable Edu-QuRaters, which can score individual text chunks on a set of educational criteria. Across two sequence-classification base models and six educational criteria, the best Edu-QuRater recovers held-out GPT-4.1-mini pairwise judgements with mean accuracy 0.917. We then apply the resulting scorers in two applications. First, we investigate the potential of Edu-QuRaters for corpus filtering to improve pretraining of small language models. We scored 322.25M FineWeb-Edu-Fortified documents to obtain a filtered pre-training mixture. In matched single-run pre-training comparisons, models trained with Edu-QuRating-based mixtures reached higher observed aggregate accuracy across nine benchmarks than the FineWeb-Edu baseline, with gains concentrated in particular tasks. Second, we used Edu-QuRater scores as reward terms for GRPO post-training. In held-out pairwise judge evaluations, combining Edu-QuRater and answer-structure rewards produced responses preferred to the Qwen3-4B base model on both pedagogical quality and instruction following.

\end{abstract}

\section{Introduction}

Data curation is now one of the central components in language-model training. Broad web filters, deduplication, and mixture design can change model performance as much as architectural choices, especially for smaller models. Educational filtering has been particularly beneficial: FineWeb-Edu shows that selecting education-oriented web text for training can improve performance on knowledge- and reasoning-heavy benchmarks relative to the same architecture trained with unfiltered web data \citep{penedo2024fineweb}. For educational applications, however, the question ``is this document educational?'' is only the beginning.

Once a corpus has already been filtered for educational content, remaining documents still vary along dimensions that matter for learning. A page can be engaging but inaccurate, accurate but badly sequenced, useful for secondary students but inaccessible to primary readers, or broadly educational without supporting foundational literacy. Treating these cases as a single educational score makes data selection hard to inspect and hard to adapt to concrete instructional goals.

Edu-QuRating takes QuRating \citep{wettig2024qurating} as its methodological starting point and specializes the approach to more fine-grained educational judgements on existing broadly educational data. Instead of asking only whether a document is educational, Edu-QuRating asks what kind of educational value it provides. QuRating showed that LLM pairwise judgements can be captured by reusable multi-dimensional single-text QuRaters, making it possible to score large corpora without querying an LLM on every document. We keep this core idea, including the pairwise preference setup and neural Bradley--Terry scoring objectives, but improve on the base model used for training the QuRaters, refine the scoring criteria toward more specific pedagogical dimensions, and introduce optimizations for gathering pairwise preferences with fewer LLM calls.

We define pairwise rubrics for factual accuracy, pedagogical structure, engagement, and learner-level suitability. We then extend these "core" educational rubrics with a set of 14 additional rubrics targeting specific components of foundational literacy education. The resulting Edu-QuRaters produce separate score dimensions that can be inspected, combined, and applied to downstream goals.

We evaluate the full path from pairwise supervision to downstream model behavior. First, we show that the pairwise LLM signal can be captured by single-text scores that accurately recover held-out pairwise preferences. Second, we characterize the training data scale and base-model choice needed for that distillation. Third, we apply the same procedure to a set of foundational-literacy rubrics. Finally, we use the learned scores for two downstream applications: filtering the web-text component of small-language-model pre-training mixtures and providing reward terms for GRPO fine-tuning of educational responses.

\section{Related Work}

\paragraph{Web-scale data curation.}
Modern pre-training recipes treat data quality as a core design choice. C4 made large Common Crawl filtering and mixture design a central part of transfer learning \citep{raffel2020exploring}; RefinedWeb and Dolma showed that documented filtering and deduplication can make open web corpora competitive and reproducible \citep{penedo2023refinedweb,soldaini2024dolma}; and DataComp-LM made curation itself a controlled benchmark \citep{li2024datacomp}. FineWeb and FineWeb-Edu are closest to our setting: FineWeb-Edu filters FineWeb into a 1.3T-token educational subset and improves several knowledge- and reasoning-heavy evaluations \citep{penedo2024fineweb}. Edu-QuRating builds on this line but changes the target of curation. Rather than asking whether a document is educational in general, we score already educational candidates along dimensions such as factual accuracy, pedagogical structure, engagement, and level suitability.

\paragraph{QuRating distills LLM judgements.}
QuRating's \citep{wettig2024qurating} core concept is to sample document pairs, ask an LLM judge for pairwise preferences, distill those preferences into QuRaters that assign scalar scores to individual documents, and then use the scores for scalable data selection. LLM-as-judge systems allow for collecting preference data that is well-aligned with human preferences \citep{zheng2023judging}. However, they do not scale well to datasets in the order of hundreds of millions of rows, due to prohibitive cost and time. The QuRating approach distills the judge's pairwise preferences from a reasonably-sized sample (in the order of hundreds of thousands of rows) into a smaller QuRater model. The methodology extends a classical paired-comparison model. In the Bradley--Terry model, each item has a latent score, and the probability of preferring one item over another is a logistic function of the score difference \citep{bradley1952rank}. Neural learning-to-rank methods such as RankNet replace fixed item scores with scores predicted from item features \citep{burges2005learning}. QuRating applies the same pairwise-logistic structure to transformer encoders: a text sequence is mapped to a scalar score in such a way that the probability of preferring one sequence over another is a logistic function of the score difference. 

The QuRater training pipeline involves a larger judge LLM supplying a set of pairwise preference labels, which are distilled into a smaller QuRater model which can predict the latent Bradley--Terry score of new documents. The trained QuRater can then be applied to new documents or generated responses at scale. Edu-QuRating keeps this structure but changes the supervision target from document quality to educational dimensions.

\paragraph{Educational criteria and foundational literacy.}
Educational usefulness is not a single property. A century of reading research models skilled reading as the product of decoding and linguistic comprehension \citep{gough1986decoding}, decomposes it into interacting word-recognition and language-comprehension strands \citep{scarborough2001connecting}, and identifies the components through which children learn to read, namely phonemic awareness, phonics, fluency, vocabulary, and comprehension \citep{nationalreadingpanel2000teaching,castles2018ending}. Synthesizing over 120 studies from low- and middle-income countries (LMICs), the GEEAP reading report adds oral language and writing to this list and reaches the conclusion that children do not learn to read incidentally: each component must be taught explicitly and systematically \citep{alvarez2025effective}.

\paragraph{Pre-training corpus}
Several recent systems show that the performance of models can be strongly influenced by their pre-training corpus. Carefully curated data can make small language models surprisingly capable, including TinyStories \citep{eldan2023tinystories}, Phi-style textbook-quality corpora \citep{gunasekar2023textbooks,li2023textbooks}, and SmolLM2 \citep{allal2025smollm2}. The Smol Training Playbook extends this line into an experimental methodology: choose a proven baseline recipe, perform small controlled ablations, and change one component at a time before scaling \citep{benallal2025smolplaybook}. Our downstream pre-training experiments follow this logic. We do not propose a new full pre-training data mixture; instead, we hold network architecture, training and implementation constant and ask whether substituting the general web-text tranche of the mixed corpus with Edu-QuRating filtered data improves the performance of an otherwise matched small-model recipe.

\paragraph{GRPO post-training}
GRPO-style reinforcement learning was introduced for mathematical reasoning in DeepSeekMath and later DeepSeek reasoning systems \citep{shao2024deepseekmath,guo2025deepseekr1}. In those settings, rewards often come from verifiable answers or task-specific correctness checks. Our setting is different: educational response quality is open-ended, so we use distilled Edu-QuRaters as learned reward models for properties such as pedagogical structure, engagement, factual accuracy, and foundational literacy support.

\section{Developing Edu-Qurater Scoring Models}

\subsection{Pipeline overview}

\begin{figure}[htbp]
    \centering
    \includegraphics[width=0.98\linewidth]{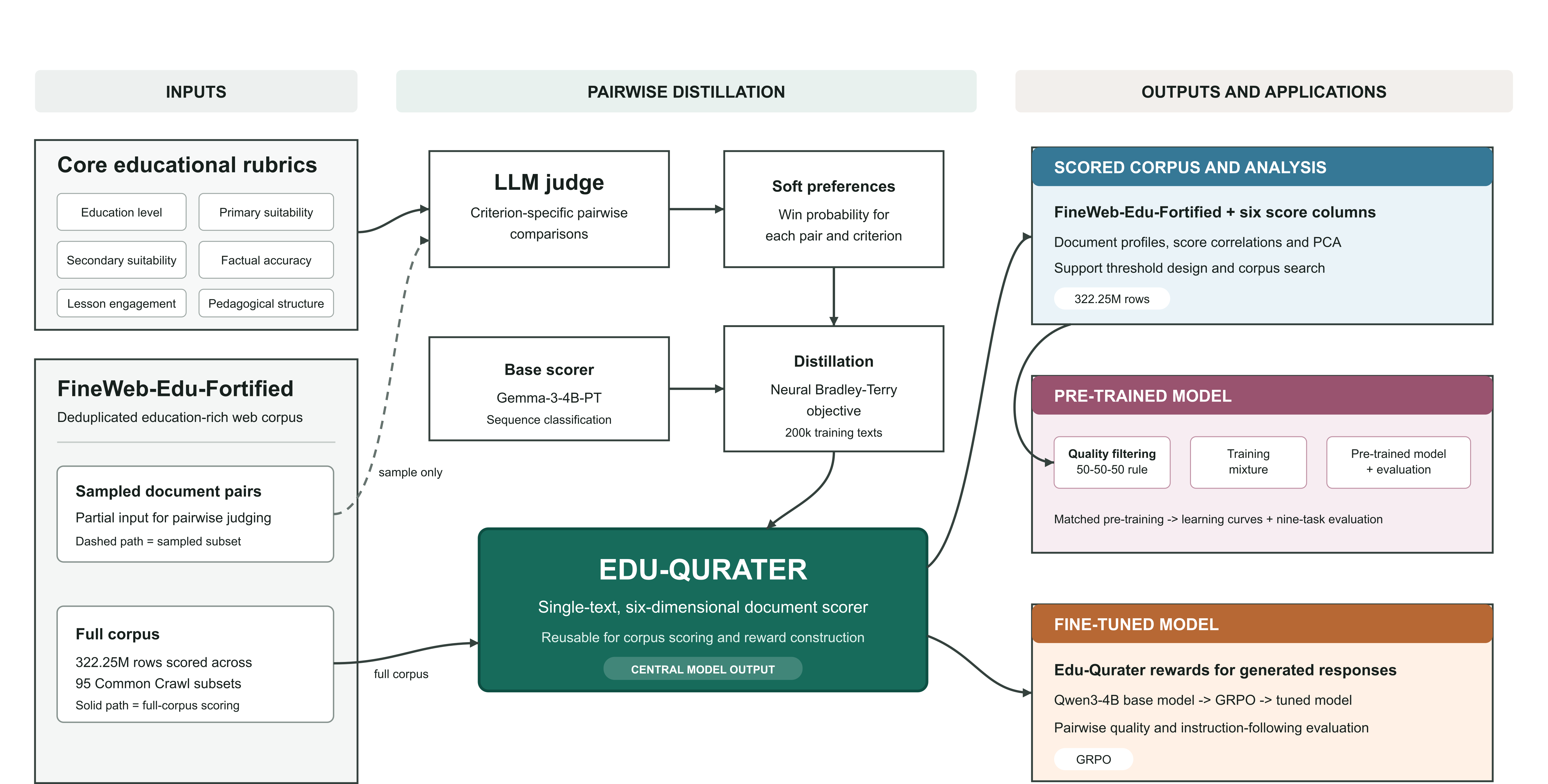}
    \caption{Edu-QuRating pipeline. Pairwise educational rubrics are used to obtain soft LLM preference labels, which are distilled into single-text Edu-QuRater score layers with a neural Bradley--Terry objective \citep{bradley1952rank,burges2005learning}. The core educational Edu-QuRater is a single multi-output model that produces six criterion-specific scores for each text. The same procedure also produces the foundational-literacy student- and teacher-facing scorers. The main comparison reports Sheared-LLaMA and Gemma-3 base models. The resulting scorers are deployed for corpus filtering, small-language-model pre-training mixtures, and response-level GRPO rewards.}
    \label{fig:edu-qurating-pipeline}
\end{figure}

Edu-QuRating adapts the QuRating preference-distillation procedure \citep{wettig2024qurating} to educational data curation. Figure~\ref{fig:edu-qurating-pipeline} illustrates the overall pipeline. First, we sample text from an education-rich web corpus. then, an LLM judge compares text pairs under criterion-specific rubrics and produces soft pairwise preference labels. A sequence-classification model is trained so that score differences between two texts predict the judge preference. Each rubric family is implemented as a single multi-output Edu-QuRater: the core educational model predicts six criterion-specific scores, while the foundational-literacy models predict their respective student-facing or teacher-facing dimensions. The trained Edu-QuRater scores the full corpus one document at a time. Finally, the scores are used either to filter pre-training data or to reward generated educational responses.

\subsection{Educational dimensions}

We define three rubric families. The core educational rubric (Core-Ed) contains six criteria: overall educational orientation, Primary-level Suitability, secondary-level suitability, factual accuracy, lesson engagement, and pedagogical structure\footnote{Core-Ed rubrics are shared here: \url{https://github.com/AI-for-Education/edu-qurating/tree/main/src/qurating/prompting/templates/ours_v2}}. These criteria are intended for general educational web text, where a document may be educational in topic but differ substantially in correctness, sequencing, accessibility, or learner engagement.

The Core-Ed set of rubrics is designed to quantify important aspects of educational quality that can vary independently across texts. A passage can be accurate but impenetrable, or engaging but wrong. Factual accuracy matters because a learner who absorbs an error is not left with a gap but with a misconception, and misconceptions can resist later correction \citep{lewandowsky2012misinformation}. Pedagogical structure rewards writing a novice can follow, with material sequenced in small steps, worked examples, and minimal extraneous detail \citep{sweller1988cognitive,rosenshine2012principles}. Lesson engagement rewards the attention hooks, relatable contexts, and response prompts that trigger and sustain a learner’s interest \citep{hidi2006fourphase}. Each dimension is judged without regard to the others, since even engaging additions can harm learning when they add extraneous load \citep{rey2012seductive}. Assumed prior knowledge and education level, by contrast, are metadata rather than quality dimensions with no universally better score. A text teaches effectively only when its demands meet the learner’s current knowledge \citep{kalyuga2003expertise}, the logic behind teaching-at-the-right-level interventions in LMICs \citep{banerjee2017proof}.

We also define two foundational-literacy rubric families. The student-facing rubric (FL-Student) targets material for beginner readers, including oral language and vocabulary, phonological awareness, systematic phonics, reading fluency, reading comprehension, writing expression, and engagement\footnote{FL-student rubrics are shared here: \url{https://github.com/AI-for-Education/edu-qurating/tree/main/src/qurating/prompting/templates/FLN_student-facing}}. The teacher-facing rubric (FL-Teacher) targets instructional material for educators, including oral-language or vocabulary instruction, phonological-awareness instruction, systematic phonics, fluency, comprehension, writing or encoding, and overall pedagogical quality\footnote{FL-teacher rubrics are shared here: \url{https://github.com/AI-for-Education/edu-qurating/tree/main/src/qurating/prompting/templates/FLN_teacher-facing}}. These two foundational-literacy rubric families are modeled directly on the GEEAP report. Each takes the report’s six components of evidence-based reading instruction and specifies what they look like in text, through two different lenses. 

The FL-Student family evaluates text as practice material, asking of each component whether a beginning reader could exercise the skill on the text itself. The oral language dimension rewards familiar, high-frequency vocabulary in culturally familiar settings, because a beginning reader can understand a written word only if they already know it from speech \citep{perfetti2014word, alvarez2025effective}. Rhyme and alliteration draw attention to the sound structures underpinning phonological awareness \citep{bryant1990rhyme}. Short, phonically regular words give beginners words they can decode, the practice through which word recognition becomes automatic \citep{ehri2014orthographic}. Repetitive frames and repeated sight words build the automaticity that frees attention for meaning \citep{laberge1974toward}, and simple, clearly sequenced sentences keep comprehension within reach of readers with limited background knowledge \citep{kintsch1988role,recht1988effect}. Prompts for oral or written response exploit the reciprocity of writing and reading \citep{graham2011writing,alvarez2025effective}. A final dimension rewards affirming, inclusive engagement, since reading growth compounds with the volume of practice that motivation sustains \citep{hidi2006fourphase,stanovich1986matthew}.

The FL-Teacher family applies the same six components to text that addresses the teacher rather than the child, asking whether it models explicit teaching practices. Following the report’s emphasis on explicit and systematic instruction \citep{alvarez2025effective}, the rubrics reward guidance that teaches word meanings and leads discussion \citep{elleman2009impact}, that practices oral blending and segmenting as distinct from print-based phonics \citep{ehri2001phonemic}, and that introduces letter-sound correspondences in sequence while penalizing cueing-based strategies that the evidence contradicts \citep{castles2018ending}. They likewise reward guided oral reading such as echo, choral, and repeated reading \citep{nationalreadingpanel2000teaching}, modeled comprehension strategies with background-knowledge building \citep{shanahan2010improving,alvarez2025effective}, and dictation and spelling tasks that link encoding back to phonics \citep{graham2012teaching}. A cross-cutting pedagogical-quality dimension rewards the delivery principles that make any component teachable. These include scaffolded modeling with gradual release of responsibility \citep{pearson1983instruction}, formative checking within the lesson \citep{rosenshine2012principles}, and alignment with the learner’s language of instruction \citep{nakamura2023language,piper2016implementing}. They are also the principles that structured-pedagogy programs, among the most cost-effective interventions in these settings, package at scale \citep{akyeampong2023costeffective}.

Table~\ref{tab:rubric-summary} summarizes these three rubric families and preference directions. Together, we define 20 output dimensions: six core educational, seven student-facing, and seven teacher-facing dimensions.

\begin{table}[!htbp]
\centering
\caption{Summary of the pairwise-comparison rubrics.}
\label{tab:rubric-summary}
\footnotesize
\setlength{\tabcolsep}{3pt}
\renewcommand{\arraystretch}{0.95}
\begin{tabular}{@{}p{0.28\linewidth}>{\scriptsize\raggedright\arraybackslash}p{0.66\linewidth}@{}}
\toprule
\textbf{Dimension} & \textbf{Summary} \\
\midrule
\multicolumn{2}{@{}l}{\textbf{Core educational quality}} \\
Educational level & Prior knowledge and conceptual difficulty, including abstraction, specialist terms, and symbolic representations. \\
Primary level & Appropriateness for primary-school learners, approximately ages 5--11. \\
Secondary level & Appropriateness for secondary-school learners, approximately ages 12--18. \\
Factual accuracy & Correctness, precision, internal consistency, and absence of unsupported claims. \\
Lesson engagement & Ability to sustain interest through relatable examples, interaction, inviting tone, and memorable imagery. \\
Pedagogical structure & Clear sequencing, accessible language, aligned examples, consistent notation, and limited irrelevant detail. \\
\midrule
\multicolumn{2}{@{}l}{\textbf{Foundational literacy: student-facing}} \\
Oral language and vocabulary & Familiar, high-frequency vocabulary and culturally familiar contexts for beginner readers. \\
Phonological awareness & Rhyme, rhythm, alliteration, and sound play that highlight spoken-word sounds. \\
Systematic phonics & Decodability through short, regular words and consistent letter--sound patterns. \\
Reading fluency & Repetitive syntax, predictable sentence stems, and repeated sight words that support fluent practice. \\
Reading comprehension & Simple sentences, coherent sequencing, questions, and picture prompts that support understanding. \\
Writing and expression & Prompts for children to answer, retell, draw, write, or otherwise respond. \\
Engagement and relevance & Affirming language, positive emotion, inclusive representation, and a child-centred tone. \\
\midrule
\multicolumn{2}{@{}l}{\textbf{Foundational literacy: teacher-facing}} \\
Oral language and vocabulary & Guidance for teaching word meanings, word parts, usage, and language comprehension. \\
Phonological awareness & Guidance for oral rhyme, blending, segmenting, syllable, onset, and rime activities. \\
Systematic phonics & Explicit, sequenced teaching of letter--sound links, decoding, blending, CVC words, and digraphs. \\
Reading fluency & Guidance for oral reading, rereading, word tracking, accuracy, pace, and expression. \\
Reading comprehension & Strategies for predicting, summarising, inferring, retelling, questioning, and building background knowledge. \\
Writing and encoding & Integration of reading, spelling, and writing through dictation, segmentation, handwriting, and sentences. \\
Pedagogical quality & Scaffolding, gradual release, formative checks, feedback, multilingual awareness, and lesson sequencing. \\
\bottomrule
\end{tabular}
\end{table}

\subsection{Pairwise preference distillation}

We used FineWeb-Edu-Fortified as our base corpus. This is an educational web corpus derived from FineWeb-Edu \citep{penedo2024fineweb}, with additional de-duplication. We sub-sampled a set of 200k documents for our pairwise LLM-judge corpus, randomly sampling proportionally from each of the 95 Common Crawl subsets which FineWeb-Edu-Fortified draws from. 


For each criterion, the judge sees two excerpts from the randomly sampled subset, together with a detailed and specific educational rubric for a specific dimension. The model is prompted to choose which of the two excerpts better satisfies the rubric criterion. Unless a criterion explicitly concerns level or audience, the prompts instruct the judge not to base the decision on text length, presentation order, or language. 

To fit in with the reduced Edu-QuRater architecture, excerpts are capped at 512 tokens. To reduce position bias, we treat comparisons as order-aware and record preferences using the forward and reverse orders. We adapted the original QuRating pipeline so that, rather than repeating many LLM-judge comparisons on the same pair of texts to build a distribution of preferences for a specific pair, we read the judge model's relative probabilities for the constrained output tokens \texttt{A} and \texttt{B} directly in a single inference pass. This approach was verified with a test run on 500 random samples. The probability sampling shortcut matched a repeated-sampling estimate with mean absolute error below 0.03 while reducing judge calls by a factor of 20. Unless otherwise specified, pairwise supervision uses \texttt{GPT-4.1-mini} and yields a soft label \(y_{ij}^{(c)} \in [0,1]\): the probability that text \(j\) is preferred to text \(i\) on criterion \(c\).

The Edu-QuRater is a sequence-classification model that outputs a score per criterion for a single input text. Its training loss is a neural Bradley--Terry objective \citep{bradley1952rank,burges2005learning}. In the classical Bradley--Terry model, each item (in this case, a document) has a scalar score, and the probability that item \(j\) is preferred to item \(i\) is determined by the logistic transform of the score difference. QuRating keeps this pairwise comparison structure but replaces fixed item scores with transformer-predicted scores. For criterion \(c\), a text \(x_i\) is assigned a score \(s_i^{(c)} = f_\theta^{(c)}(x_i)\). Given scores \(s_i^{(c)}\) and \(s_j^{(c)}\), the implied preference probability is
\[
    \hat{p}_{ij}^{(c)} = \sigma\left(s_j^{(c)} - s_i^{(c)}\right),
\]
where \(\sigma\) is the logistic sigmoid. Training minimizes binary cross-entropy between \(\hat{p}_{ij}^{(c)}\) and the LLM-derived soft label \(y_{ij}^{(c)}\). This objective distills pairwise supervision into single-text scores that can later be evaluated independently.

Unlike the traditional Bradley-Terry model, the model does not learn a separate parameter for each document seen during training. Instead, documents provide token sequences, and the encoder is trained end-to-end to map those sequences to criterion-specific scores.

We compare two sequence-classification base models: Sheared-LLaMA-1.3B (following the original QuRating implementation) and Gemma-3-4B-PT. To test the effect of data scale on model performance, we fix the base model to Sheared-LLaMA-1.3B and vary the number of pairwise examples. The base model comparison holds the number of training examples constant at 200k, uses 512-token excerpts, trains for two epochs, and uses a 0.1 warmup fraction. For held-out evaluation we score text pairs independently and check whether the score difference recovers the \texttt{GPT-4.1-mini} pairwise preference \footnote{Edu-Qurating models are available at: \url{https://huggingface.co/collections/AI-for-Education/edu-qurating-models}}.

\subsection{Edu-Quraters accurately distill pairwise educational comparisons}

We first test whether the trained Edu-Qurating models accurately predict pairwise LLM preferences through the scores they assign to individual texts. We evaluate trained Edu-QuRaters on held-out \texttt{GPT-4.1-mini} comparisons, with 50,000 evaluated pairs per criterion. Each Edu-QuRater scores the two texts independently; the sign of the score difference determines the implied pairwise winner.

\begin{figure}[htbp]
    \centering
    \includegraphics[width=1.0\linewidth]{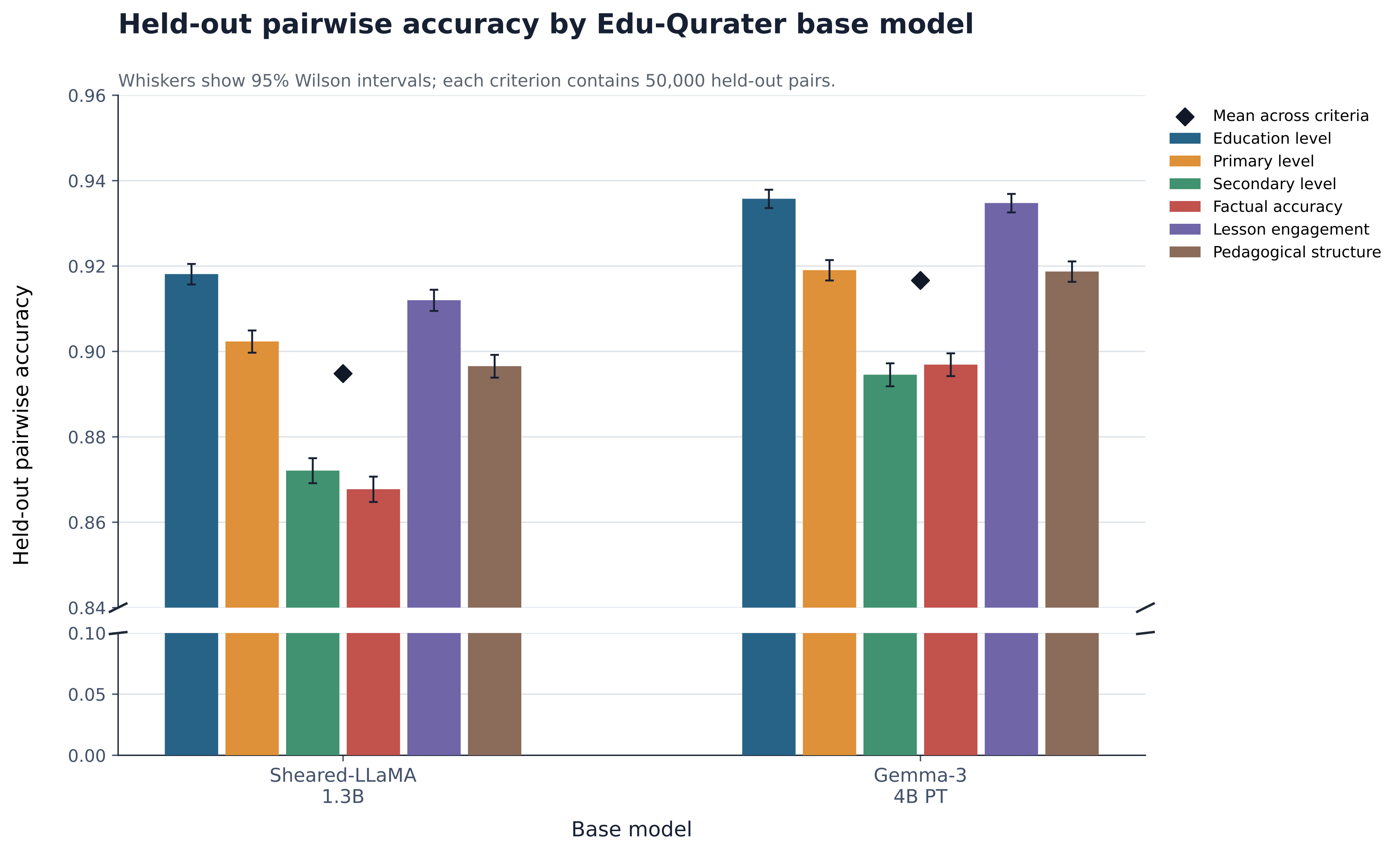}
    \caption{Held-out pairwise accuracy by Edu-QuRater base model. Colored bars show the six educational criteria, with two-sided 95\% Wilson score intervals computed over 50,000 held-out pairs per criterion; black diamonds show the unweighted mean across criteria. Pairwise preferences are provided by \texttt{GPT-4.1-mini}.}
    \label{fig:qurater-model}
\end{figure}

Across the two base models and six criteria, Edu-QuRaters recover the held-out pairwise preferences with high accuracy (Figure~\ref{fig:qurater-model}). Both base models exceed 0.86 accuracy on every criterion. Mean accuracy rises from 0.895 for Sheared-LLaMA-1.3B to 0.917 for Gemma-3-4B-PT. Gemma improves the accuracy uniformly across every criterion.

This evaluation test more than just reproducing the LLM judge preferences with another pairwise model. At inference time, the Edu-QuRater does not jointly inspect the two texts and return a preference. It produces single-text scores, and pairwise preferences are recovered only from score differences. This is the key property of the pipeline: once pairwise educational preferences are distilled, a single scorer can be applied to new documents or model-generated responses without rerunning the LLM judge.

\subsection{Scaling of distillation accuracy with number of pairwise judgements}

We next examine the amount of pairwise supervision needed to achieve a good distillation result. This test holds the Sheared-LLaMA-1.3B base model constant and varies the number of pairwise examples used for training. Performance is measured by validation loss on held-out pairwise probabilities; lower loss means that the model's predicted preference probabilities more closely match the LLM-derived labels. For each run, we report the best validation loss observed across evaluation checkpoints.

\begin{figure}[htbp]
    \centering
    \includegraphics[width=0.9\linewidth]{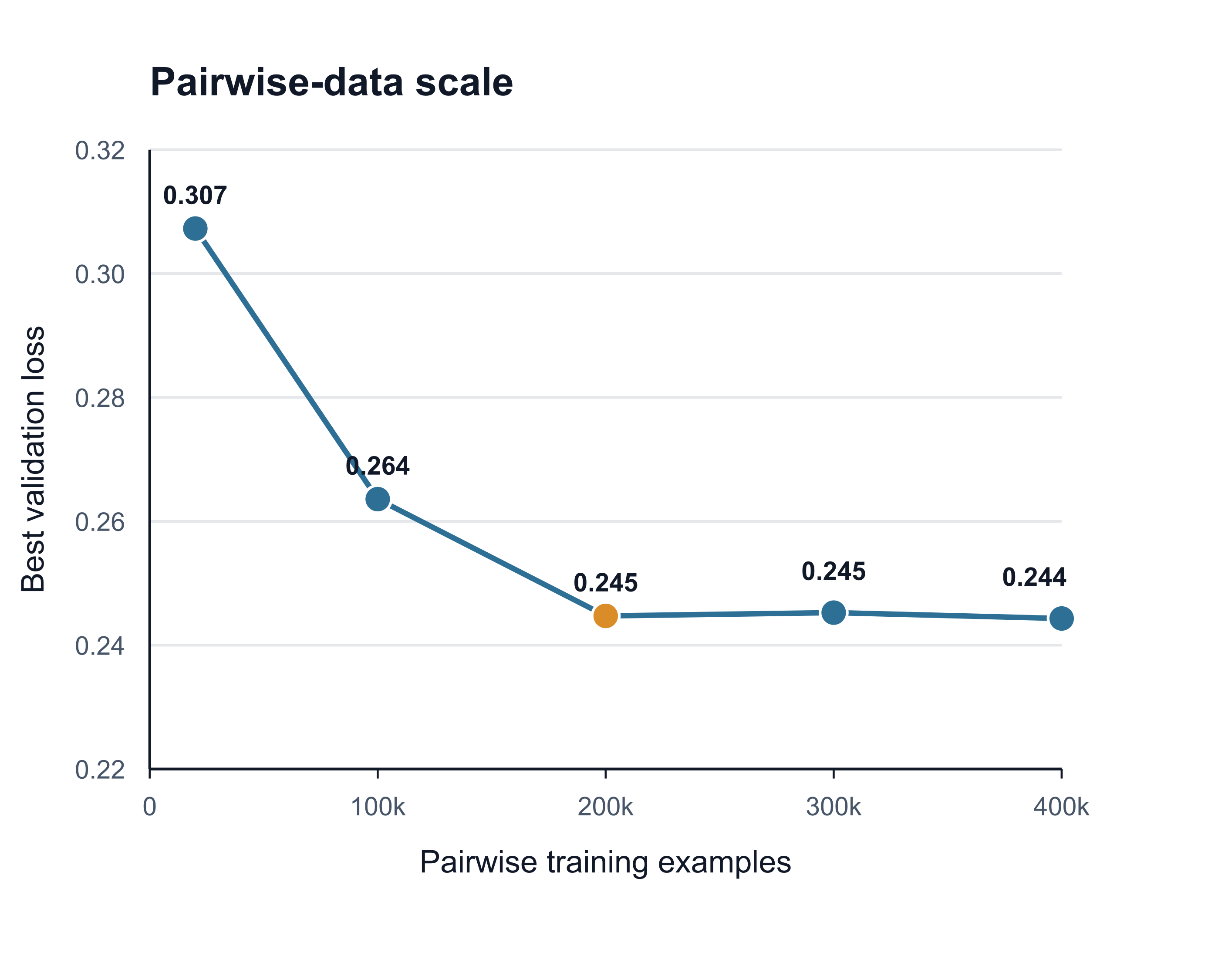}
    \caption{Pairwise-data scale. Best validation loss observed across evaluation checkpoints for Sheared-LLaMA-1.3B Edu-QuRaters trained with different numbers of \texttt{GPT-4.1-mini} pairwise examples.}
    \label{fig:distillation-ablation}
\end{figure}

Validation loss decreases sharply from 20k to 200k examples, falling from 0.307 at 20k to 0.264 at 100k and 0.245 at 200k (Figure~\ref{fig:distillation-ablation}). Increasing to 300k or 400k examples yields almost no additional gain, with best losses of 0.245 and 0.244. For this baseline base model, 200k examples therefore provide a practical training scale: the run reaches the low-loss region observed in larger runs without requiring substantially more LLM-labelled comparisons.

Together, Figure~\ref{fig:qurater-model} shows that trained Edu-QuRaters recover held-out pairwise preferences from single-text scores, while Figure~\ref{fig:distillation-ablation} identifies 200k pairwise examples as a practical data scale for the baseline setup. The web-scale scoring results below use the Gemma-3-4B-PT Edu-QuRater.

\subsection{Foundational Literacy Edu-QuRaters}

The preceding results use general educational-quality criteria. We next ask whether the same pairwise-distillation procedure can be applied to a more specific educational goal: scoring of materials specifically focused on learning foundational-literacy. Following the same pipeline, we train two literacy-facing Edu-QuRater families. The student-facing scorer (FL-Student) targets material for learners developing literacy skills; the teacher-facing scorer (FL-Teacher) targets instructional material for educators and is also used later as part of the GRPO reward configurations.

The pipeline worked very similarly here. Gemma-3-4B-PT reached validation losses of 0.128 for student-facing literacy scoring and 0.175 for teacher-facing literacy scoring. As a qualitative sanity check, applying the core and literacy Edu-QuRaters to external educational materials yields structured variation by scraped education level and material type, suggesting that the retargeted scorers are not merely reproducing a single generic educational-quality signal. More details on that analysis follows.

\subsection{Distribution of Edu-QuRater scores on out-of-sample educational materials}

Before applying Edu-Quraters to downstream applications, we inspect whether score profiles vary sensibly across education level and material type when applied to a dataset of external educational materials \citep{ccbottomup2026}. These materials are separate from the FineWeb-Edu-Fortified corpus used for Edu-Qurater distillation and web-scale filtering. While not a direct test, it provides an indirect indicator of the validity of the Edu-Qurating models when scoring unseen materials.

The full external dataset consists of \textasciitilde{50,000} materials. We selected \textasciitilde{15,800} materials from this where we had reliable standardized annotations for material type and education level and we were able to estimate the language as English. For language estimation we used the provided language annotation where available, and otherwise fell back to estimates using the \texttt{lingua} python package.


\begin{figure}[htbp]
    \centering
    \includegraphics[width=\linewidth]{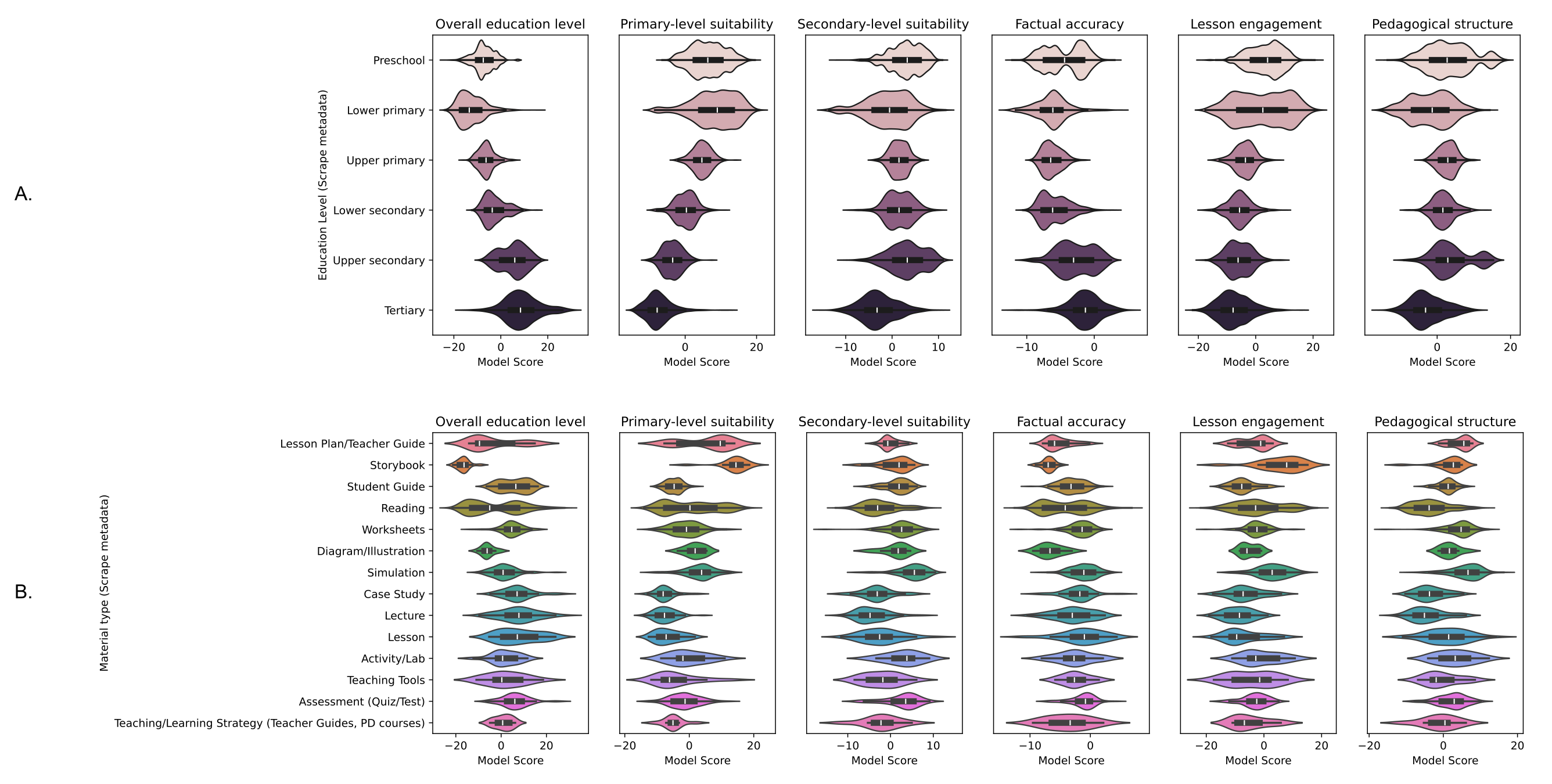}
    \caption{Score distributions from Core-Educational Edu-Qurater model over 15k sourced educational materials divided by metadata tags. (A) Distributions divided by Education Level metadata; (B) Distributions divided by Material type metadata.}
    \label{fig:distr-core-ed}
\end{figure}

\begin{figure}[htbp]
    \centering
    \includegraphics[width=\linewidth]{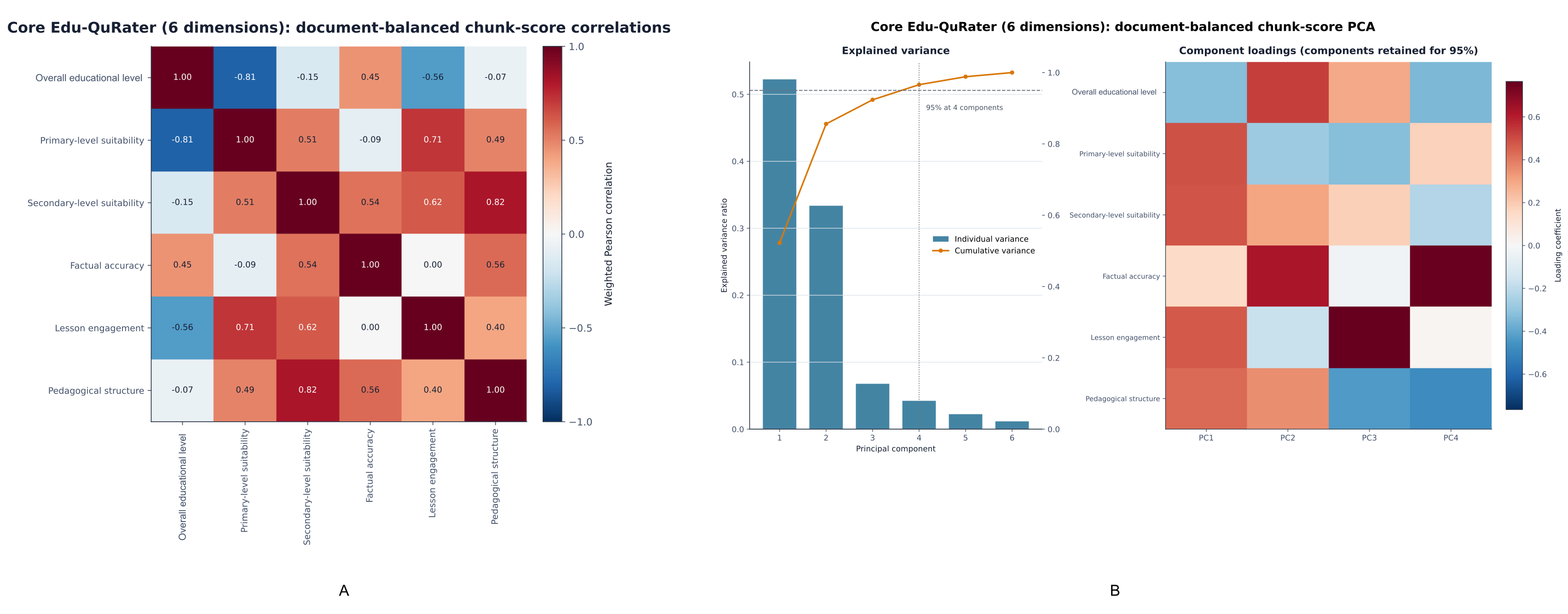}
    \caption{(A) Correlation matrix and (B) PCA between document-weighted chunk scores from the six dimensions of Core-Educational Edu-Qurater over 15k sourced educational materials.}
    \label{fig:core-ed-corr}
\end{figure}

Figure~\ref{fig:distr-core-ed} shows the distributions of Edu-Qurating scores from the 6 Core-Ed dimensions in the out-of-sample educational materials. The Core-Ed dimensions vary sensibly when split by the materials' education level metadata (Figure~\ref{fig:distr-core-ed}(A)): notably the \textbf{Overall Education Level} dimension appears highly correlated with the education level metadata and the \textbf{Primary-level Suitability} displays a peak around the primary education levels of the metadata. Marginal Pearson correlations between education level metadata and Core-Ed dimensions are highest for these two dimensions (0.62 and 0.64 respectively). A linear model of education level metadata which combines all six dimensions achieved a cross-validated \(R^2\) of 0.449, compared with 0.387 (\textbf{Overall Education Level}) and 0.405 (\textbf{Primary-level Suitability}) for the strongest two single-dimension models. 

Another interesting feature is the apparent bimodal distribution of factual accuracy scores in pre-school materials. This could reflect increased use of fantastical or fairy-tale content to capture students attention, while not being factually accurate as described in our rubric. 

Figure \ref{fig:core-ed-corr} shows the inter-dimension correlation and PCA between the 6 score distributions over the full \textasciitilde15k materials sample. \textbf{Overall Education Level} displays a strong negative correlation with both \textbf{Primary-level Suitability} and \textbf{Lesson Engagement}, which are both strongly correlated with each other. Four principal components explain 95\% of the variance. The largest component contrasts \textbf{Overall Education Level} with the remaining dimensions; the second largest component contrasts \textbf{Primary-level Suitability} and, to a lesser extent,  \textbf{Lesson Engagement} with the other dimensions (strongest contrast on \textbf{Factual accuracy} and \textbf{Overall Education Level}).

The corresponding FL-Student and FL-Teacher profiles, together with correlations across all 20 dimensions, are reported in Appendix~\ref{appendix:score_profiles}

\FloatBarrier

\section{Application: Improving small model performance with an Edu-QuRater filtered pre-training corpus}

\subsection{Methods}

\subsubsection{Corpus Scoring and Filtering}\label{sec:score_filter}
\paragraph{Scoring}
For corpus-scale scoring, we apply the Gemma-3-4B-PT Edu-QuRater to FineWeb-Edu-Fortified. Documents are tokenized, split into 512-token chunks, scored chunk by chunk, and reduced to document-level scores by averaging over chunks. The scoring run covers all 95 Common Crawl subsets and produces scores for 322.25M rows.

For scalability, we use an orchestrator / worker pipeline for scoring, where the orchestrator is responsible for first dividing the 95 Common Crawl subsets into batches of roughly 100,000 rows then, for each batch, instantiating an H200 worker VM and launching the above scoring process on the worker with the corresponding subset indices to score (the worker itself is responsible for streaming the data from each subset in each batch as well as scoring each row). Scoring is performed on 32 H200 GPUs. Scoring the entire dataset of 322.25M rows takes \textasciitilde72 hours (\textasciitilde2300 GPU hours in total), with each single GPU scoring at a rate of \textasciitilde40 rows per second on average.

\paragraph{Filtering}
After scoring, we filter the training corpus by thresholding on percentile values across scores. For downstream pre-training experiments we filter only on the three "quality" dimensions of the Core-Ed Edu-Qurater model: \textbf{Factual Accuracy}, \textbf{Lesson Engagement}, and \textbf{Pedagogical Structure}. We view the three educational level dimensions -- \textbf{Overall Education Level}, \textbf{Primary-level Suitability}, \textbf{Secondary-level Suitability} -- as metadata labels which can be used for targeting specific audiences but do not necessarily provide a quality signal in the absence of additional context.

We selected a conjunctive 50-50-50 rule, where each ``50'' denotes the 50th-percentile threshold for \textbf{Factual Accuracy}, \textbf{Lesson Engagement}, and \textbf{Pedagogical Structure}, respectively. That is, we keep only materials that score above the median level of the full dataset in all 3 quality dimensions. The resulting corpus contains 51.94M rows, or 16.12\% of the 322.25M scored rows; the retained fraction is below 50\% because the three thresholds are applied conjunctively. In addition to the 50-50-50 rule condition, during one of our runs we inadvertently sampled with a stricter threshold for pedagogical structure and more forgiving thresholds for factual accuracy and lesson engagement (exact percentiles are 41.6 for factual accuracy, 34.4 for lesson engagement, and 69.6 for pedagogical structure). We include this in our experiments as a separate "stricter pedagogy" condition. 

\subsubsection{Pre-training}
To test whether the Edu-QuRating-filtered corpus improves downstream evaluations when used for model pre-training, we chose to follow an established pre-training pipeline and architecture. The Smol Training Playbook \citep{benallal2025smolplaybook} provides us with an established baseline and recipe into which we insert the Edu-QuRating-filtered dataset. Rather than attempt to replicate the full-scale pre-training run reported in the Playbook, we instead replicate the much smaller 1B parameter ablation configuration\footnote{Baseline nanotron configuration is here: \url{https://huggingface.co/datasets/HuggingFaceTB/training-guide-nanotron-configs/blob/main/baseline_config_1B.yaml}} which comprises training runs over 45B tokens from a mixture of three datasets. We train on a cluster of 8 H100 GPUs using the nanotron framework and each run takes approximately 48 hours.

Following the Smol Training Playbook baseline configuration, all data mixtures keep 10\% FineMath-3Plus and 20\% Stack-Edu-Python fixed. For our experiments, the remaining 70\% web-text component is either FineWeb-Edu, Edu-QuRating 50-50-50 filtered, the stricter-pedagogy Edu-QuRating filtered dataset, or a 35\%/35\% mixture of the stricter-pedagogy dataset and DataComp-LM (DCLM), a general-purpose web corpus \citep{li2024datacomp}. We train matched small language models and evaluate saved checkpoints with a subset of the LightEval suite: Pedagogy CDPK cloze, ARC-CF, MMLU-CF, MMLU-Pro-CF, BoolQ-CF, CommonsenseQA-CF, OpenBookQA-CF, HellaSwag, and WinoGrande.

\subsection{Results}

\subsubsection{Edu-Qurating Scores Meaningfully Span the Educational Space}

To test whether the six Core-Ed dimensions recover signal captured by an established scalar filter while retaining distinct information, we compare their document-level scores with the FineWeb-Edu score across FineWeb-Edu-Fortified. Marginal Pearson correlations are modest, ranging from 0.042 for lesson engagement to 0.326 for \textbf{Secondary-level Suitability}; no individual dimension therefore closely reproduces the FineWeb-Edu score. A linear model combining all six dimensions achieves a cross-validated \(R^2\) of 0.228, compared with 0.106 for the strongest single-dimension model. Leave-one-predictor-out results show the largest reduction after removing \textbf{Secondary-level Suitability} (\(R^2=0.139\)), while \textbf{Overall Education Level} also contributes complementary multivariate information despite its weak marginal correlation. Otherwise, single dimensions add relatively small amounts of unique predictive information once the other dimensions are included, suggesting overlap among some criteria. Together, these results indicate that Edu-QuRater captures part of the educational-quality signal represented by FineWeb-Edu while decomposing it into dimensions that are not individually redundant with the scalar score.

\subsubsection{Edu-QuRating-Filtered Data Reaches Higher Observed Endpoints in Matched Pre-Training}

The downstream experiments compare the observed learning curves of models trained on Edu-QuRating-selected corpora under a matched training pipeline. We follow the ablation setup from the Smol Training Playbook in which the non-web components remain constant across conditions: every mixture contains 10\% FineMath-3Plus and 20\% Stack-Edu-Python. The only manipulated component is the remaining 70\% web-text slice. The baseline uses FineWeb-Edu for this slice, whereas the Edu-QuRating conditions replace it with filtered educational text, either alone or mixed with DCLM.

\begin{table}[htbp]
    \centering
    \small
    \begin{tabular}{p{0.18\linewidth}p{0.30\linewidth}p{0.14\linewidth}p{0.14\linewidth}}
        \toprule[1.5pt]
        \centering{Mixture} & \centering{Web / educational text} & \centering{Math} & \centering{Code} \tabularnewline
        \toprule[1.5pt]
        \centering{FineWeb-Edu baseline} & \centering{70\% FineWeb-Edu} & \centering{10\% FineMath-3Plus} & \centering{20\% Stack-Edu-Python} \tabularnewline
        \midrule[0.5pt]
        \centering{Edu-QuRating 50-50-50} & \centering{70\% regenerated Edu-QuRating-filtered data} & \centering{10\% FineMath-3Plus} & \centering{20\% Stack-Edu-Python} \tabularnewline
        \midrule[0.5pt]
        \centering{Edu-QuRating stricter pedagogy} & \centering{70\% stricter-pedagogy Edu-QuRating data} & \centering{10\% FineMath-3Plus} & \centering{20\% Stack-Edu-Python} \tabularnewline
        \midrule[0.5pt]
        \centering{Edu-QuRating stricter pedagogy + DCLM} & \centering{35\% stricter-pedagogy Edu-QuRating data + 35\% DCLM} & \centering{10\% FineMath-3Plus} & \centering{20\% Stack-Edu-Python} \tabularnewline
        \bottomrule
    \end{tabular}
    \newline
    \caption{Pre-training mixtures used in the downstream comparison. The stricter-pedagogy artifact comes from the audited earlier filtering output and is reported separately from the regenerated intended 50-50-50 corpus.}
    \label{tab:pretraining-mixtures}
\end{table}

At saved pre-training checkpoints, each model is evaluated with the same LightEval suite: Pedagogy CDPK cloze, ARC-CF, MMLU-CF, MMLU-Pro-CF, BoolQ-CF, CommonsenseQA-CF, OpenBookQA-CF, HellaSwag, and WinoGrande. We report mean accuracy across all nine tasks, as well as the mean over the five education-related tasks, and the four general tasks separately. The education-related mean averages Pedagogy CDPK cloze, ARC-CF, MMLU-CF, MMLU-Pro-CF, and OpenBookQA-CF; the general-task mean averages BoolQ-CF, CommonsenseQA-CF, HellaSwag, and WinoGrande.

\begin{table}[htbp]
    \centering
    \small
    \begin{tabular}{c|c}
        \toprule[2pt]
        Educational Tasks & Description \\
        \midrule
        ARC-CF & Grade school science reasoning\\
        MMLU-CF & Broad academic knowledge across 57 subjects\\
        MMLU-Pro & Challenging academic knowledge across 14 subjects \\
        OpenBookQA & Elementary science facts with reasoning \\
        Pedagogy & Pedagogical knowledge from teacher exams\\
        \toprule[2pt]
        General tasks & Description\\
        \midrule
        BoolQ-CF & Naturally-occurring yes/no questions paired with a short web passage\\
        CommonsenseQA & Common-sense reasoning about everyday concepts\\
        HellaSwag & Common-sense reasoning about everyday situations (narrative completion)\\
        WinoGrande & Pronoun resolution requiring world knowledge\\
        \bottomrule
    \end{tabular}
    \newline
    \caption{Division of benchmarks by Educational and General tasks. Descriptions taken from \citep{benallal2025smolplaybook} where applicable}
\end{table}

\begin{figure}[htbp]
    \centering
    \includegraphics[width=1.0\linewidth]{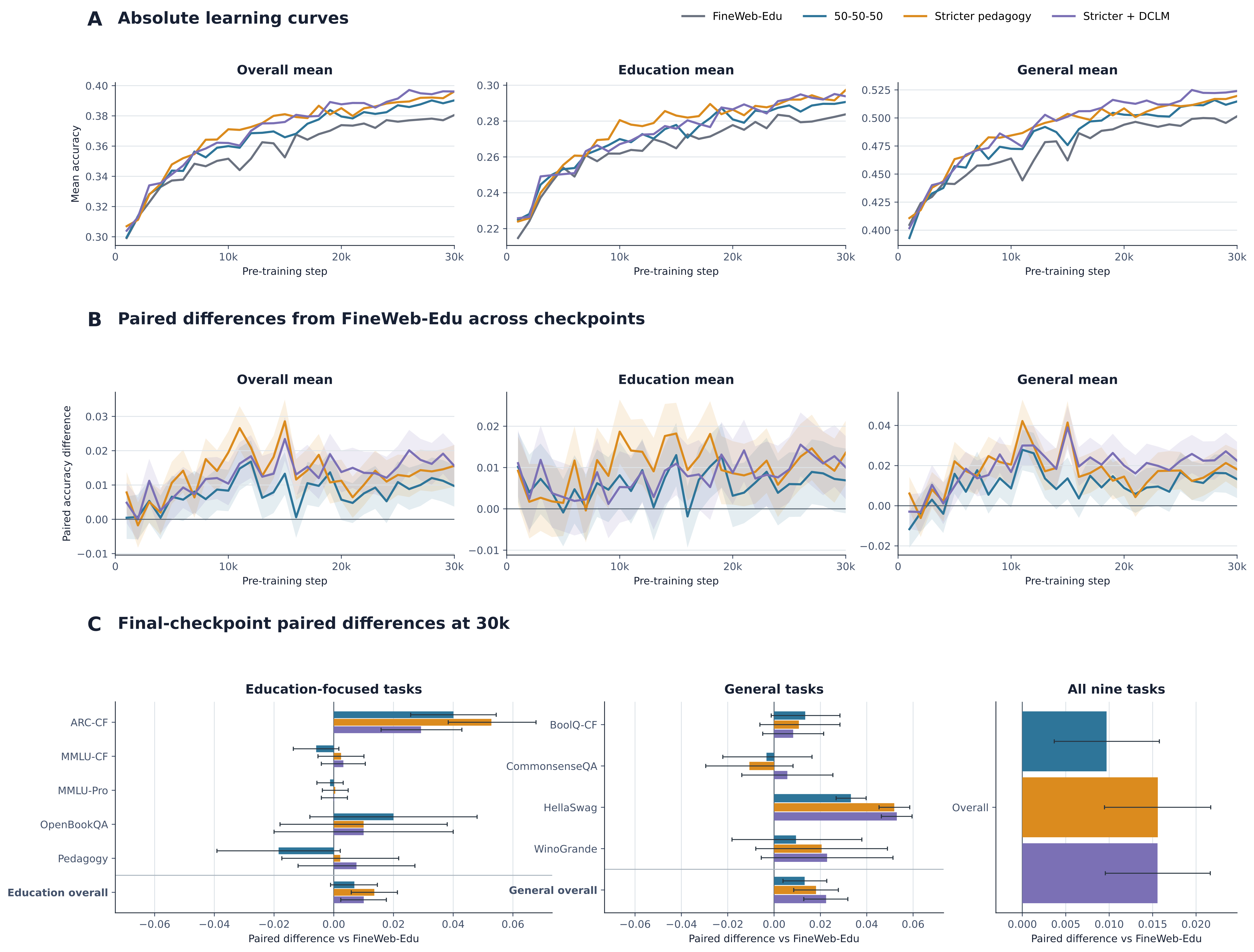}
    \caption{Downstream LightEval results by pre-training mixture. (A) Learning curves show mean accuracy across all nine tasks, the education-related subset, and the general-task subset through the 30k-step endpoint. 
    (B) Curves show gains relative to FineWeb-Edu baseline across all nine tasks, the education-focused subset, and the general-task subset through the 30k-step endpoint. (C) Final-checkpoint gains relative to the FineWeb-Edu baseline are shown for each benchmark, the education-related and general-task means, and the overall nine-task mean; positive values indicate higher accuracy than the matched baseline. Shaded bands and whiskers show 95\% percentile intervals from 10,000 paired item-level bootstrap replicates. The same resampled items are used across conditions and checkpoints; ARC is stratified by challenge/easy subset and MMLU by subject. These intervals capture evaluation-set uncertainty conditional on the trained checkpoints, not training-seed variation.}
    \label{fig:slm-downstream}
\end{figure}

At 30k steps, all Edu-QuRating-based mixtures reach higher observed aggregate means than the FineWeb-Edu baseline in matched single-run comparisons (Figure~\ref{fig:slm-downstream}(A)). The baseline reaches 0.3806 mean accuracy across the nine tasks. The regenerated 50-50-50 Edu-QuRating run reaches 0.3903, a gain of 0.0097 over the baseline. The stricter-pedagogy run reaches 0.3962, and the Edu-QuRating + DCLM mixture reaches 0.3961. These two conditions are effectively tied on the aggregate endpoint, but they differ in profile: the stricter-pedagogy Edu-QuRating-only run has the highest education-related mean, while the mixed DCLM run has the highest general-task mean.

The 50-50-50 condition tests the intended filtering rule, and its observed 30k endpoints are higher than FineWeb-Edu on the overall, education-related, and general-task means. The stricter-pedagogy condition gives the strongest Edu-QuRating-only result and the strongest education-related endpoint. The mixed Edu-QuRating + DCLM condition gives the strongest general-task mean, suggesting that Edu-QuRating-filtered educational text and high-quality general web text may complement each other.

The endpoint view at task-level shows that the gain is concentrated in a few specific evaluation tasks (Figure~\ref{fig:slm-downstream}(C)). ARC-CF and HellaSwag show the clearest improvements, with gains of roughly three to five accuracy points depending on the Edu-QuRating mixture. OpenBookQA, BoolQ, and WinoGrande also improve for all Edu-QuRating conditions, though by smaller margins. In contrast, the MMLU-family task improvements are close to zero, CommonsenseQA is mixed, and Pedagogy CDPK cloze depends on the filtering variant. The evidence therefore supports a measured conclusion: Edu-QuRating-filtered mixtures reach higher observed aggregate endpoints and higher scores on several individual tasks in these matched single-run comparisons, but the differences are task-specific rather than uniform across benchmarks.

\section{Application: Edu-Quraters as GRPO Reward Models}

\subsection{Methods}

\subsubsection{Fine-tuning: GRPO}

We wanted to investigate whether Edu-QuRaters can help to improve generated educational responses through direct feedback during fine-tuning rather than filtering data for pre-training. As well as being a more direct application of the scorers, this is also a more accessible application: as outlined in \ref{sec:score_filter}, scoring the full 322M rows of the pre-training corpus costs considerable compute resources, before considering the resources required for the pre-training run, whereas GRPO fine-tuning can achieve results with training data on the order of thousands of rows\citep{chen2026dragrpogrponeedsknow,dang2026reinforcementlearningreasoningsmall}.

To prompt the model, we use a proprietary dataset that contains a set of teacher tasks focused on foundational literacy for grades 0-3 along with examples of high quality outputs. The training set contains 1008 examples of teacher tasks and AI-generated high-quality outputs from a large frontier model (\texttt{gemini-3.1-pro}). In each GRPO run, the base model receives an educational prompt, generates candidate responses, and receives a score from a reward function. The tested reward combinations include Edu-Qurating model scores -- the Core-Ed quality dimensions such as pedagogical structure, lesson engagement, and factual accuracy, and the FL-Teacher dimensions -- and a separate auxiliary reward for matching answer-structure.

\paragraph{Reward functions}
We first test the GRPO procedure with only the answer-structure reward function. The reward function operates with an LLM-judge scoring outputs against the synthetic high-quality output data outlined above. Outputs receive one point for matching the length of the high quality response, one point for matching the overall formatting style, and one point for following the prompt instructions to the same degree as the high quality response. For the judge we use \texttt{gemma-4-E4B} running on a local \texttt{llama.cpp} server.

For the second test, we want to see if Edu-Qurating scores alone can improve response quality without the cost of generating this synthetic dataset and with increased scalability. We implement parallel reward functions for the Core-Ed and FL-Teacher Edu-Quraters. For the Core-Ed reward function, we average the scores for Factual Accuracy, Lesson Engagement, and Pedagogical Structure. For the FL-Teacher reward function, we average the scores from all 7 dimensions.

Finally, we combine both approaches -- reference against high quality output and Edu-Qurating reward models. As above, this is implemented as parallel reward functions within the same configuration. We also vary here the contribution of different Edu-Qurater models -- either Core-Ed and FL-Teacher combined, or Core-Ed only.

We train on a single RTX6000 with 48GB VRAM using the \texttt{unsloth} framework\citep{unsloth}, with the Edu-Qurating models and the \texttt{gemma-4-E4B judge served by llama.cpp} running on the same GPU in parallel. We train for a total of 2000 steps per run, sampling with replacement from the 1008 training examples.

\paragraph{Evaluation}
The setup separates the reward optimization from the final evaluation. GRPO updates the model toward responses with higher rewards, but the reported metric is a held-out pairwise judge comparison between GRPO checkpoint outputs and the Qwen3-4B base model. We evaluate both overall quality of educational response and instruction-following. This is important because a reward model can make a response more polished or pedagogically structured but fail to answer the user's request. 

The evaluation dataset comprises 73 held-out examples with task instructions similar to those in the training set, supplemented by additional human quality control from education experts. In addition, each item in the dataset is tagged for relevance to each of the 7 FL-Teacher dimensions, with multiple tags possible per item.

The pedagogical quality prompt consists of a base instruction along with additional FL-Teacher-specific instructions, which are appended on a per-item basis for each corresponding tag that item has. The base instruction asks the judge to consider: age-appropriateness for the stated grade; accuracy and subject-matter correctness; concreteness and practical classroom usefulness; clarity for the target teacher audience; cultural / contextual fit for LMIC classrooms; with explicit instructions to ignore format and length. The specific FL-Teacher instructions adopt the same rubric criteria as the Edu-Qurater rubrics.

The instruction following prompt asks the judge to consider: required counts ("3 objectives", "5 activities"); length caps ("each under 12 words", "one paragraph"); required format / structure (bullets, numbered list, table, sections); target grade / age group; target subject or topic; required elements (action verbs, specific vocabulary, etc.); with explicit instructions to ignore pedagogical quality, usefulness, accuracy, or style.

Each response pair is judged in both order configurations (base mode response first or fine-tuned model response first) and final reported win rates are aggregated over both configurations. The judge model is \texttt{gemini-3-flash-preview}.

\subsection{Results}

\subsubsection{Edu-QuRater Rewards Improve Educational Responses after GRPO}

\begin{figure}[htbp]
    \centering
    \includegraphics[width=\linewidth]{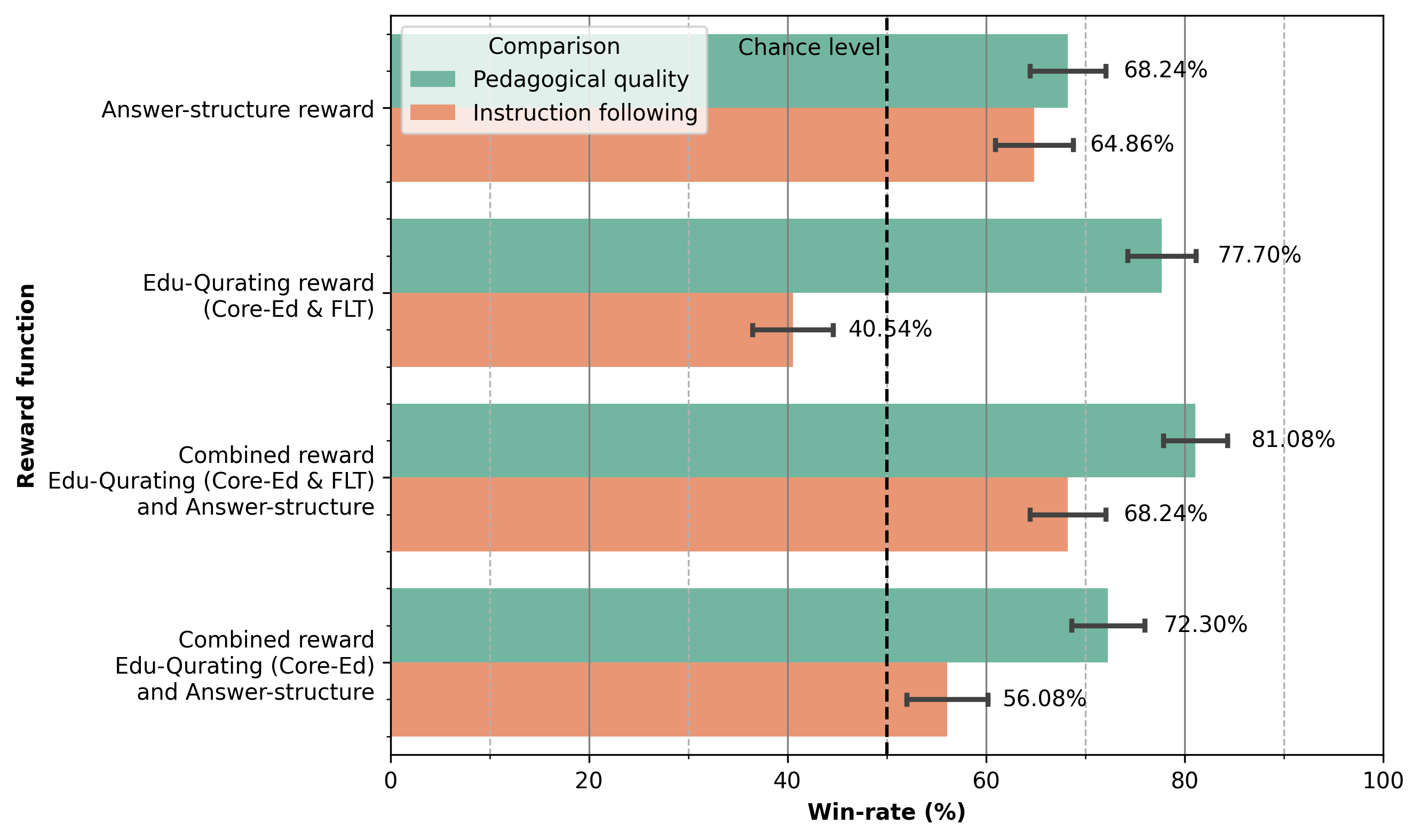}
    \caption{Win rates against the Qwen3-4B base model on pedagogical quality and instruction following for four GRPO reward configurations. Bars show pairwise-judge preference rates and error bars show standard error. The dashed line marks 50\% parity with the base model.}
    \label{fig:grpo-main-comp}
\end{figure}

Figure~\ref{fig:grpo-main-comp} summarizes the main results. The answer-structure reward alone produces win rates greater than 50\% compared to the base model in both pedagogical quality (68.24\%) and instruction following (64.86\%). This condition validates the overall GRPO procedure and provides a baseline for the subsequent reward configurations. Although the reward targets length, formatting, and adherence to instruction rather than content directly, these structural properties can themselves contribute to perceived pedagogical quality.

The Edu-QuRating reward alone produces the highest pedagogical-quality win rate of the two single-source reward conditions (77.70\%), but its instruction-following win rate falls below parity with the base model (40.54\%). This pattern is consistent with over-optimizing a proxy reward that does not directly constrain prompt adherence, although the present evaluation does not establish the mechanism.

Combining the Edu-QuRating and answer-structure rewards produces the highest observed pedagogical-quality win rate (81.08\%) while retaining an instruction-following win rate above parity with the base model (68.24\%).

\FloatBarrier

\section{Discussion}

Edu-QuRating adapts pairwise preference distillation to educational data curation. We introduced rubrics for general educational quality and student- and teacher-facing foundational literacy, examined how base-model choice and supervision scale affect distillation, and applied the resulting scorers to corpus filtering and GRPO reward modeling\footnote{Source code for gathering pairwise preferences; Edu-QuRater training, scoring, and filtering; and GRPO training with Edu-QuRater rewards is available at: \url{https://github.com/AI-for-Education/edu-qurating}.}\footnote{Our nanotron fork, including run configurations and data preparation scripts, is available at: \url{https://github.com/AI-for-Education/nanotron}.}. High agreement with held-out judge preferences shows that these criteria can be practically applied at scale. The downstream experiments demonstrate their utility in selecting pre-training material from an already educational corpus and providing feedback on generated educational responses.

Our pre- and post-training results show that Edu-Qurater scoring can directly benefit model training. All tested Edu-QuRating pre-training mixtures achieved higher observed aggregate accuracy than the FineWeb-Edu baseline. For GRPO fine-tuning, combining Edu-QuRater and answer-structure rewards improved both judged pedagogical quality and instruction following. Edu-QuRater rewards can be computed without reference answers, making them potentially useful where high-quality examples are costly to produce.

These educational model training demonstrations are relevant to educational products in a global context, including low- and middle-income countries, where content needs to suit local curricula, languages of instruction, and classroom conditions. But our approach also supports educational content curation applications more broadly, through both quality assessment (which was used for the model-training applications) and scalable metadata labeling. Scores for factual accuracy, pedagogical structure, and engagement can guide quality-based selection, while learner-level suitability and foundational-literacy scores can help describe the audience and instructional purpose of material. The student- and teacher-facing literacy rubrics offer an example of curation around specific instructional uses. Adapting the scoring criteria with educators familiar with local curricula and classroom conditions could make the resulting scores more useful for content selection in specific contexts. With validated mappings from scores to labels, Edu-QuRating could support indexing, retrieval, recommendation, and the assembly of training or evaluation datasets for a wide range of educational products. The key point is that Edu-QuRaters can help systematic content curation at scale by provided relevant educational metadata which is often not available from the data source. 

Several limitations constrain these findings, however. The scorers inherit the coverage of the upstream corpus and cannot compensate for missing languages, curricula, or genres. Due to resource constraints, the pre-training experiments use one run per mixture and we were unable to run a more comprehensive set of mixture combinations. The experiment therefore cannot claim to establish the most effective possible mixture, only that our stricter-pedagogy plus DCLM mixture performed the best out of those which were tested. In addition, the GRPO evaluation is relatively small scale and the pedagogical criteria for the model evaluation were designed to overlap with the FL-Teacher Edu- QuRating rubrics. The degree to which the model improvements from GRPO generalise beyond the FL-Teacher framework defined here was not tested explicitly. Nonetheless, we consider both of these experiments successful proofs-of-concept which demonstrate the potential broader utility of the Edu-Qurating framework. 

Evaluation of Edu-QuRater impact in real-world usage of educational products would be important to demonstrate that the observed model improvements translate into genuine classroom benefits. Educator assessments of selected materials and generated responses, alongside studies of learner outcomes, would provide evidence of their practical value across teaching contexts. However, this was beyond the scope of this proof-of-concept technical report. 

Edu-QuRating provides a method for educational curation at scale, supporting both quality assessment and metadata labeling. Its use in pre-training and reward modeling shows how explicit educational criteria can guide model development, with broader potential to help teams organize and select content for the learners and teachers their products serve.

\bibliographystyle{plainnat}
\bibliography{references}

\appendix
\setcounter{figure}{0}
\setcounter{table}{0}
\renewcommand{\thefigure}{\thesection.\arabic{figure}}
\renewcommand{\thetable}{\thesection.\arabic{table}}
\renewcommand{\theHfigure}{\thesection.\arabic{figure}}
\renewcommand{\theHtable}{\thesection.\arabic{table}}


\section{External Score Profiles}\label{appendix:score_profiles}

\begin{figure}[htbp]
    \centering
    \includegraphics[width=\linewidth]{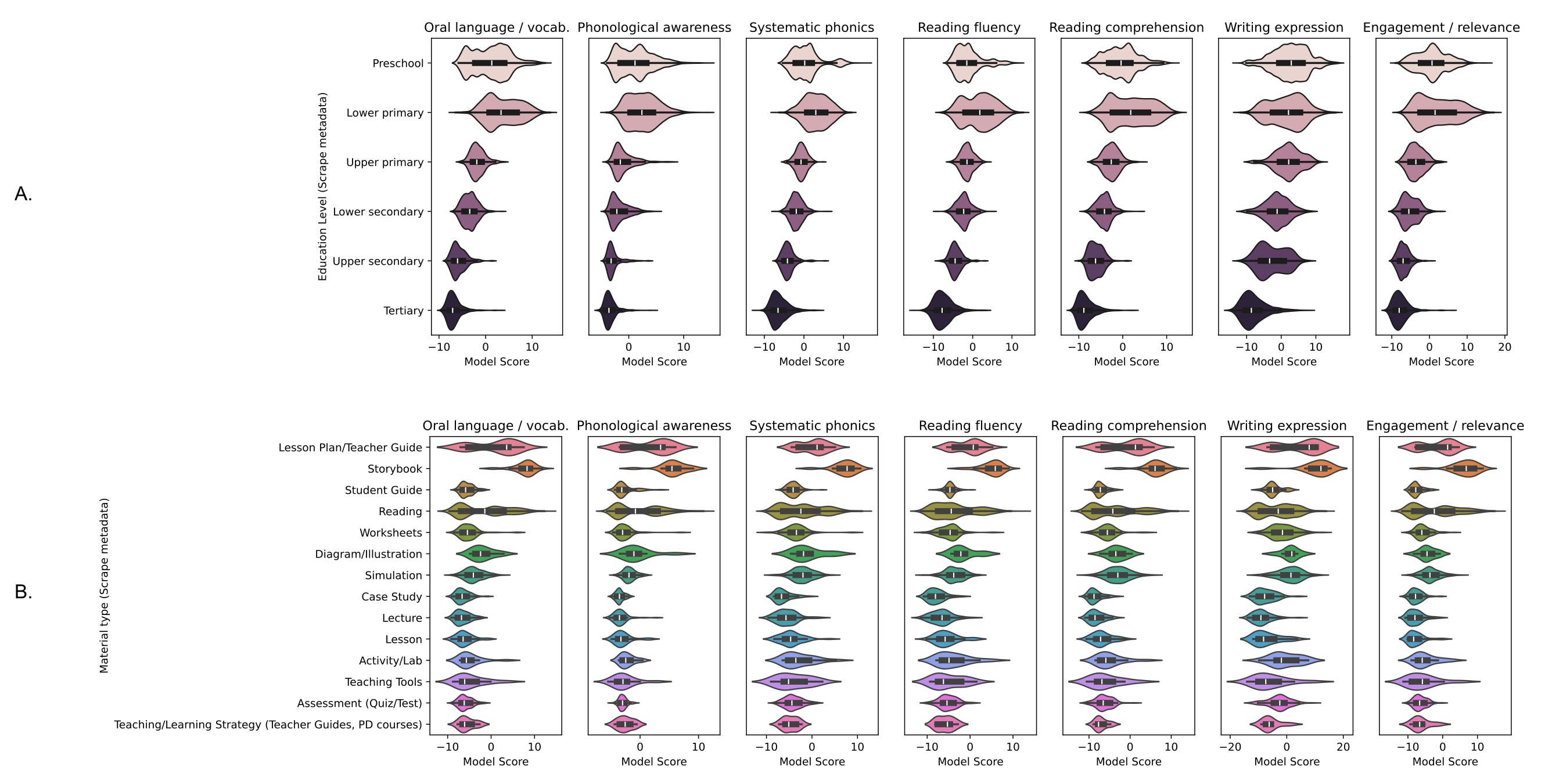}
    \caption{Score distributions from FL-Student Edu-Qurater model over 15k sourced educational materials divided by metadata tags. (A) Distributions divided by Education Level metadata; (B) Distributions divided by Material type metadata.}
    \label{fig:distr-fl-student}
\end{figure}

\begin{figure}[htbp]
    \centering
    \includegraphics[width=\linewidth]{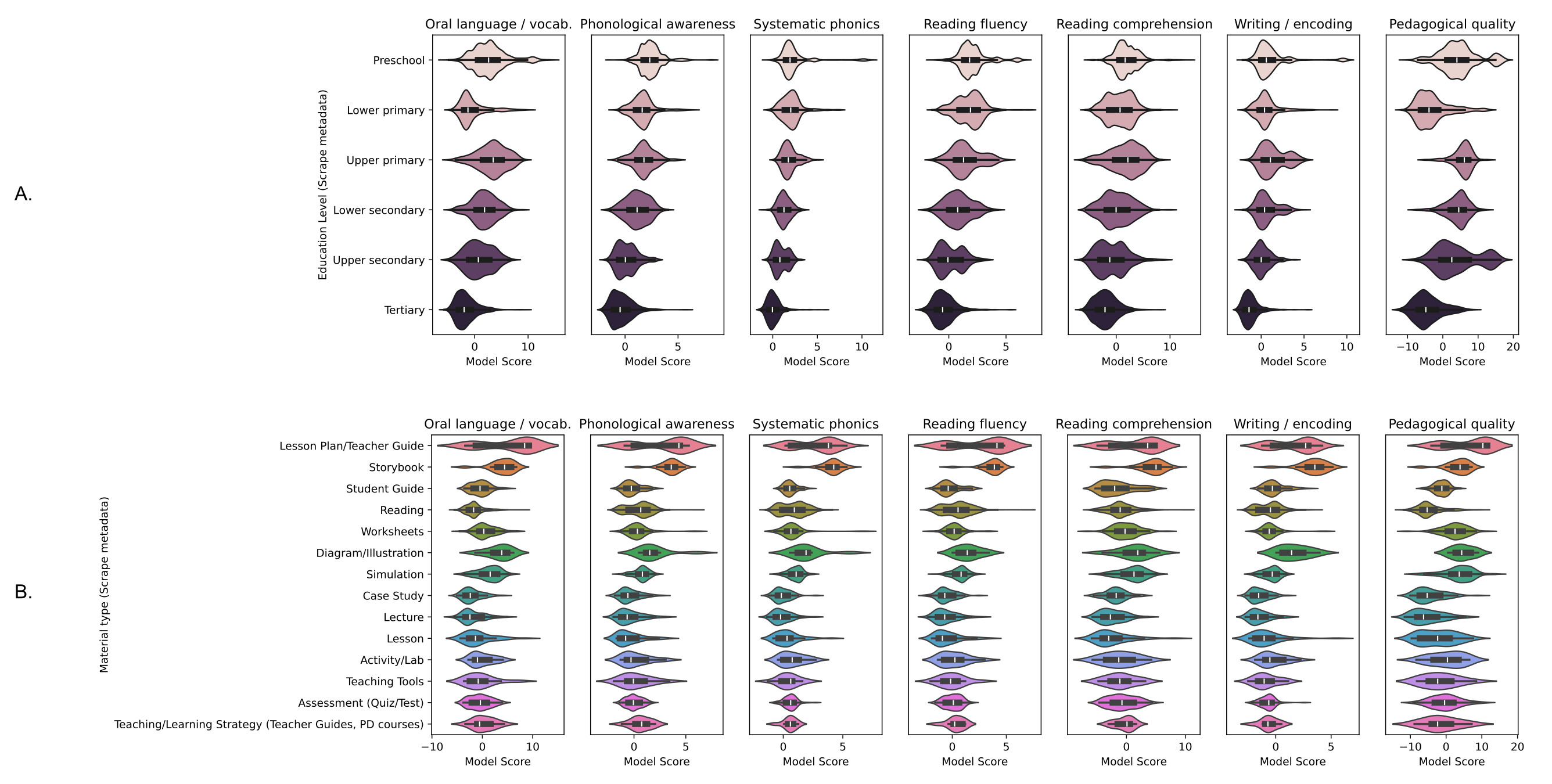}
    \caption{Score distributions from FL-Teacher Edu-Qurater model over 15k sourced educational materials divided by metadata tags. (A) Distributions divided by Education Level metadata; (B) Distributions divided by Material type metadata.}
    \label{fig:distr-fl-teacher}
\end{figure}

\begin{figure}[htbp]
    \centering
    \includegraphics[width=\linewidth]{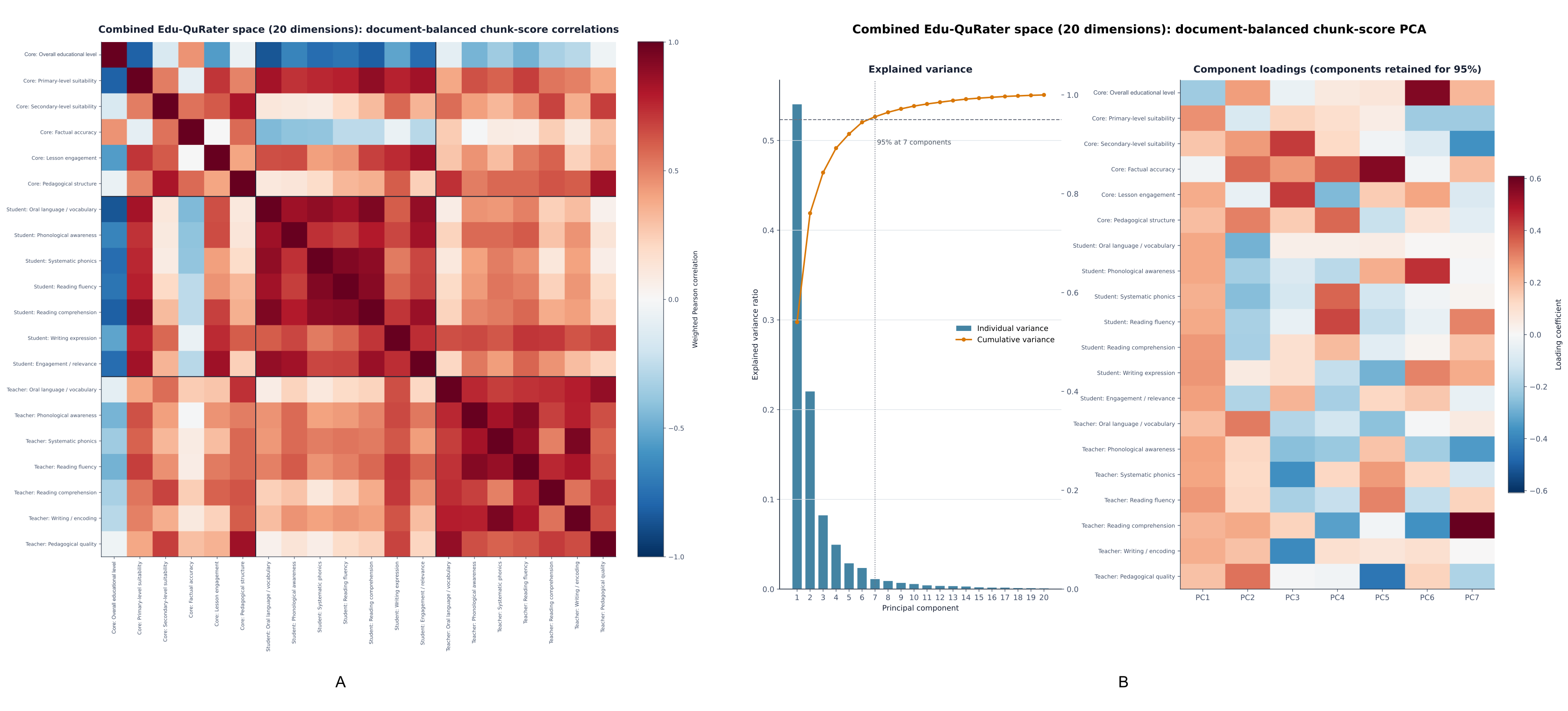}
    \caption{(A) Correlation matrix and (B) PCA between document-weighted chunk scores from the twenty combined dimensions of Core-Educational, Student-, and Teacher-Facing Foundational Literacy Edu-Quraters over 15k sourced educational materials.}
    \label{fig:full-20-corr}
\end{figure}



\end{document}